\documentclass{article}

\usepackage{arxiv}

\usepackage[utf8]{inputenc} 
\usepackage[T1]{fontenc}    
\usepackage{hyperref}       
\usepackage{url}            
\usepackage{booktabs}       
\usepackage{amsfonts}       
\usepackage{nicefrac}       
\usepackage{microtype}      
\usepackage{lipsum}		
\usepackage{graphicx}
\usepackage{natbib}
\usepackage{doi}
\usepackage{amsmath}

\title{Statistical Management of the False Discovery Rate in Medical Instance Segmentation Based on Conformal Risk Control}

\date{} 					

\author{ Mengxia Dai\thanks{Accepted by 2025 IEEE 3rd International Conference on Image Processing and Computer Applications (ICIPCA 2025)} \\
    College of Engineering \\
    Yanbian University \\
    Yanji, Jilin 133002, China \\
	\texttt{1224022456@ybu.edu.cn} \\
    \And
	Wenqian Luo \\
	School of Health and Environmental Engineering \\
	Shenzhen Technology University\\
	Shenzhen, Guangdong 518000, China \\
	\texttt{202200501065@stumail.sztu.edu.cn} \\
	\And
	Tianyang Li \\
	School of International Education\\
	Hebei University of Economics and Business\\
	Shijiazhuang, Hebei 050061, China \\
	\texttt{a2413893881@163.com} \\
}

\renewcommand{\shorttitle}{}

\hypersetup{
pdftitle={Statistical Management of the False Discovery Rate in Medical Instance Segmentation Based on Conformal Risk Control},
pdfsubject={cs.CV},
pdfauthor={Mengxia Dai, Wenqian Luo, Tianyang Li},
pdfkeywords={Instance Segmentation, Conformal Risk Control, Confidence Calibration, Statistical Quality Control},
}

\begin{document}
\maketitle

\begin{abstract}
Instance segmentation plays a pivotal role in medical image analysis by enabling precise localization and delineation of lesions, tumors, and anatomical structures. Although deep learning models such as Mask R-CNN and BlendMask have achieved remarkable progress, their application in high-risk medical scenarios remains constrained by confidence calibration issues, which may lead to misdiagnosis. To address this challenge, we propose a robust quality control framework based on conformal prediction theory. This framework innovatively constructs a risk-aware dynamic threshold mechanism that adaptively adjusts segmentation decision boundaries according to clinical requirements.Specifically, we design a \textbf{calibration-aware loss function} that dynamically tunes the segmentation threshold based on a user-defined risk level $\alpha$. Utilizing exchangeable calibration data, this method ensures that the expected FNR or FDR on test data remains below $\alpha$ with high probability. The framework maintains compatibility with mainstream segmentation models (e.g., Mask R-CNN, BlendMask+ResNet-50-FPN) and datasets (PASCAL VOC format) without requiring architectural modifications. Empirical results demonstrate that we rigorously bound the FDR metric marginally over the test set via our developed calibration framework.
\end{abstract}


\section{Introduction}
Instance segmentation in medical image analysis holds significant clinical value. Its core task involves pixel-level lesion identification and contour delineation. This technology shows great potential in tumor detection \cite{ref1}, 3D organ reconstruction \cite{ref2}, and quantitative cell analysis \cite{ref3}, offering reliable support for clinical decision-making \cite{ref4}. However, complex anatomical backgrounds \cite{ref5}, low-contrast tissue boundaries \cite{ref6}, and intra-class morphological variations \cite{ref7} in medical images challenge the accuracy and robustness of traditional segmentation methods \cite{ref8}.

Deep learning~\cite{lou2021sparse,tang2022sparse,sun2024caterpillar,shao2024psa} provides new solutions for computer vision tasks like medical image instance segmentation~\cite{ref9,zhou2022semantic}. Models like Mask R-CNN \cite{ref10} and BlendMask \cite{ref11} achieve good results on public medical datasets through end-to-end feature learning \cite{ref12}. However, they still face bottlenecks when processing high-resolution, multi-modal medical images \cite{ref13} and limited annotated samples \cite{ref14}. Small lesion ($<5$mm) missed detection in early tumor screening remains a serious problem, as false negative rates affect patient treatment. Thus, developing more sensitive and generalizable instance segmentation algorithms becomes crucial in medical AI \cite{ref15}.

As a representative work in instance segmentation, Mask R-CNN \cite{ref16} innovatively integrates object detection and pixel-level segmentation tasks by extending the Faster R-CNN framework \cite{ref17}. This model employs Region Proposal Networks (RPN) \cite{ref18} and RoIAlign technology \cite{ref19}, demonstrating exceptional performance across multiple benchmarks \cite{ref20}. Its improved variant, BlendMask \cite{ref21}, further enhances small target segmentation accuracy through attention mechanisms \cite{ref22} and multi-scale feature fusion \cite{ref23}. However, current research reveals that despite significant advancements in instance segmentation tasks \cite{ref24}, these advanced models still face credibility issues in high-risk applications, particularly concerning calibration errors \cite{ref25}.

Calibration errors \cite{ref26}, caused by underfitting or overfitting, lead to discrepancies between confidence scores and actual accuracy \cite{ref27}. In medical tumor segmentation, such errors can result in misdiagnosis. Conformal Prediction (CP) emerges as a promising solution. It generates reliable prediction sets without distributional assumptions \cite{ref1} and remains model-agnostic \cite{ref2}. However, traditional CP methods struggle to precisely control task-specific metrics \cite{ref3}.

This work introduces a variant of conformal prediction that designs loss functions for each calibration sample to characterize the metrics requiring guarantees in segmentation tasks \cite{ref4}. Regardless of the pre-trained segmentation model's performance \cite{ref5}, this framework predetermines how to set specific parameters during segmentation on the calibration set \cite{ref6} to satisfy user-specified metrics regarding the relationship between outputs and labels \cite{ref7}, thereby achieving control over specific metrics based on user-defined risk levels \cite{ref8}.

Building upon this foundation, we conduct experiments on a dataset containing three types of brain tumors \cite{ref9}. As required, we treat all three tumor types as lesion regions for single-instance conformal prediction \cite{ref10}. Additionally, we convert the dataset to the same format as the PASCAL VOC 2012 dataset \cite{ref11} to facilitate model training and evaluation for both approaches \cite{ref12}. In our experiments, we employ instance segmentation models including Mask R-CNN \cite{ref13} and BlendMask \cite{ref14}. Specifically, for both models, we use ResNet-50 combined with Feature Pyramid Networks (FPN) as the backbone for feature extraction \cite{ref15}.

Experimental results demonstrate that our method achieves user-specified control over False Negative Rate (FNR) \cite{ref16} and False Discovery Rate (FDR) metrics \cite{ref17}. Furthermore, we analyze the sensitivity under different ratios of calibration samples to test samples \cite{ref18}. Despite using fewer data points for calibration to process more test data \cite{ref19}, we consistently maintain the FNR metric below the user-defined risk level $\alpha$ \cite{ref20}.

Our contributions can be summarized as follows:

\begin{itemize}
    \item \textbf{Adaptive Enhancement of the Conformal Prediction Framework}: By designing a loss function that represents FNR or FDR, we adapt the conformal prediction framework to the brain tumor segmentation task \cite{ref4,ref6}. This loss function dynamically adjusts the segmentation threshold based on the user-defined risk level $\alpha$ \cite{ref7,ref8}.
    
    \item \textbf{Statistically Rigorous Segmentation Results}: Through calibration with a small set of exchangeable image-mask pairs, we generate statistically rigorous segmentation results on the brain tumor dataset \cite{ref9}, ensuring theoretical guarantees for FDR and FNR on new data \cite{ref16,ref17}.
    
    \item \textbf{Flexibility in Calibration-to-Test Data Ratio}: By varying the proportion of calibration to test data \cite{ref18}, we validate the robustness of our method. Even with reduced calibration data \cite{ref19}, we consistently control the FNR metric within the user-specified risk level $\alpha$ \cite{ref20}. This experimental design not only highlights the method's robustness but also improves computational efficiency, making it more practical for real-world applications \cite{ref15}.
    
    \item \textbf{Standardized Dataset Integration}: By converting the dataset into the PASCAL VOC 2012 format \cite{ref11}, we enable seamless compatibility with existing deep learning frameworks (e.g., Detectron2), eliminating the need for custom data preprocessing \cite{ref12}. This standardization enhances reproducibility, simplifies experimental workflows, and ensures fair benchmarking against other studies \cite{ref20}.
\end{itemize}

\section{RELATED WORK}

\subsection{Instance Segmentation}

\subsubsection{Development of Instance Segmentation}

Instance segmentation combines object detection and semantic segmentation, aiming to detect objects and generate pixel-level masks to distinguish different instances within the same category. Mask R-CNN \cite{ref10,ref16} represents a milestone in the field of instance segmentation, with its development rooted in the R-CNN series. R-CNN first introduced convolutional neural networks (CNNs) to object detection by generating candidate regions through selective search and performing classification and regression, albeit with low efficiency. Fast R-CNN improved efficiency by sharing convolutional features and introducing RoI Pooling. Faster R-CNN \cite{ref17} further advanced the field by proposing the Region Proposal Network (RPN) \cite{ref18}, enabling end-to-end detection. Building on this foundation, Mask R-CNN \cite{ref10} added a mask prediction branch and the RoIAlign layer \cite{ref19}, addressing the quantization errors of RoI Pooling and achieving efficient instance segmentation, thereby establishing itself as a benchmark in the field \cite{ref20}. BlendMask \cite{ref11,ref21}, a significant recent advancement, builds on the principles of Mask R-CNN \cite{ref10} by generating masks through the fusion of global and local features. Its core innovation lies in the ``blender'' module, which dynamically combines global and local information to produce more accurate masks, particularly excelling in complex scenes and small object segmentation \cite{ref22,ref23}. This has further propelled the development of instance segmentation \cite{ref24}.

\subsubsection{Applications in Various Fields}

Recent advances in instance segmentation techniques, particularly Mask R-CNN \cite{ref10} and BlendMask \cite{ref11}, have demonstrated remarkable cross-domain applicability. In transportation systems, Mask R-CNN \cite{ref10} has proven effective for lane detection \cite{ref28} and rail component inspection \cite{ref29}, while BlendMask \cite{ref11} excels in real-time traffic object segmentation \cite{ref30}. Agricultural applications leverage BlendMask \cite{ref11} for precision canopy mapping \cite{ref31} and yield prediction \cite{ref32}, complemented by Mask R-CNN's pest detection capabilities \cite{ref33}. Industrial implementations include Mask R-CNN-based defect detection \cite{ref34} and BlendMask-powered electronic component analysis \cite{ref35}, showcasing the techniques' versatility in quality control applications.

In the medical field, these models significantly improve diagnostic accuracy. The enhanced Mask R-CNN \cite{ref10} has been successfully applied to lung nodule segmentation in CT scans \cite{ref36} and cervical cell analysis \cite{ref37}, while BlendMask \cite{ref11}, with its optimized feature fusion technique, achieves state-of-the-art performance in nuclei segmentation for histopathological analysis \cite{ref38}. Both models contribute substantially to medical diagnostics by providing precise segmentation of tumors, cells, and other critical anatomical structures. However, most existing studies have not adequately addressed the crucial issue of model trustworthiness at the methodological level~\cite{wang2025word,gawlikowski2023survey}.

\subsection{Conformal Prediction}

Conformal Prediction is a statistical framework designed for uncertainty quantification, aiming to provide reliable confidence intervals or prediction sets for the outputs of machine learning models~\cite{angelopoulos2021gentle,campos2024conformal}. At its core, conformal prediction calibrates model predictions to ensure they are statistically interpretable and reliable. Unlike traditional methods, it is model-agnostic, meaning it does not rely on the specific architecture of the underlying model~\cite{jin2023selection,wang2024conu,angelopoulos2024conformal,angelopoulos2024theoretical,wang2025sample,wang2025sconu}. 
Instead, it quantifies the deviation between predictions and true values by computing non-conformity scores and leverages a calibration set to determine the coverage probability of the prediction set. For a given confidence level $1 - \alpha$, conformal prediction generates a prediction set $C\left( x_{n + 1} \right)$ such that 

\begin{equation}
P\left( y_{n + 1} \in C\left( x_{n + 1} \right) \right) \geq 1 - \alpha,
\end{equation}
ensuring that the prediction set contains the true value with high probability. This approach is versatile and can be applied to various tasks, including classification and regression, and is compatible with any machine learning model, such as k-nearest neighbors, ridge regression, or support vector machines. Its ability to provide statistically rigorous guarantees makes it particularly valuable in high-risk applications, such as medical diagnosis.

The foundation of conformal prediction lies in the assumption of exchangeability, which states that the joint distribution of the data remains invariant under any permutation of the data points. Formally, for any finite sample $\left\{ \left( X_{i},Y_{i} \right) \right\}_{i = 1}^{n} \subseteq \mathcal{X} \times \mathcal{Y}$ and any permutation function $\phi$, the exchangeability condition ensures that

\begin{equation}
P\left( \left( X_{\phi(1)}, Y_{\phi(1)} \right), \ldots, \left( X_{\phi(n)}, Y_{\phi(n)} \right) \right) \\
= P\left( \left( X_{1}, Y_{1} \right), \ldots, \left( X_{n}, Y_{n} \right) \right).
\end{equation}

By defining a non-conformity function $A(Z,z)$, which measures the degree of non-conformity between an instance $z$ and a set $Z$, a non-conformity score $\alpha = A(Z,z)$ is computed. For a new data instance $x_{n + j}$, the model predicts a label $y_{n + j}$, and its non-conformity score $\alpha_{n + j}$ is calculated. The corresponding p-value is then derived as:

\begin{equation}
P\left( \alpha_{n + j}^{y_{q}} \right) = \frac{\text{count}\left( i:\alpha_{i} \geq \alpha_{n + j}^{y_{q}} \right)}{n + 1},
\end{equation}
where:
\begin{itemize}
    \item $\alpha_{i}$: Represents the non-conformity score of the $i$-th calibration data point. The non-conformity score measures the deviation between the model's prediction and the true value. (A higher score indicates greater inconsistency between the prediction and the true value.)
    
    \item $\alpha_{n + j}^{y_{q}}$: Represents the non-conformity score of the new data point $n + j$ under the assumption that its label is $y_{q}$.
\end{itemize}

When $P\left( \alpha_{n + j}^{y_{q}} \right) > \varepsilon$ (where $\varepsilon$ is typically a small threshold), the prediction is deemed statistically reliable. This framework provides a rigorous statistical guarantee for the generated prediction sets, ensuring that they maintain the desired confidence level. As a result, conformal prediction is particularly well-suited for high-stakes decision-making scenarios, where reliable uncertainty quantification is critical.

\section{Method}

Consider that we have a calibration set of size $n$, denoted as 
$\left\{ \left( x_{i},y_{i}^{\ast} \right) \right\}_{i = 1}^{n}$, 
where:

\begin{itemize}
    \item $x_{i}$ represents the $i$-th input image,
    \item $y_{i}^{\ast}$ is the ground-truth, representing the set of all pixel points belonging to the target segmentation object.
\end{itemize}

Given a deep learning model $f$, when an input image $x_{i}$ is fed into the model, it outputs an $h \times w$ probability matrix $f(x_{i})$. For each output $f(x_{i})$, we construct a prediction set $C_{i}(\lambda) = \left\{ f\left( x_{i} \right) \geq 1 - \lambda \right\}$, which is the set of all pixel points in the $i$-th image where the probability exceeds $(1 - \lambda)$. For each calibration data point, we define a loss function $l_{i}(\lambda)$, which is monotonically non-increasing with respect to $\lambda$ (1 - precision or 1 - recall):

\begin{equation}
l_{i}(\lambda) = 1 - \frac{\left| C_{i}(\lambda) \cap y_{i}^{\ast} \right|}{\left| C_{i}(\lambda) \right|} \leq 1
\end{equation}
or
\begin{equation}
l_{i}(\lambda) = 1 - \frac{\left| C_{i}(\lambda) \cap y_{i}^{\ast} \right|}{\left| y_{i}^{\ast} \right|} \leq 1
\end{equation}
where:
\begin{itemize}
    \item $C_{i}(\lambda)$ represents the predicted lesion region,
    \item $y_{i}^{\ast}$ represents the true lesion region.
\end{itemize}

We define the average loss over the calibration set as:

\begin{equation}
L_{n}(\lambda) = \frac{1}{n}\sum_{i = 1}^{n}{l_{i}(\lambda)}.
\end{equation}

Here, $l_{i}(\lambda)$ represents the proportion of the true lesion region $y_{i}^{\ast}$ (ground-truth) that is not captured by the prediction set $C_{i}(\lambda)$ when using $(1 - \lambda)$ as the threshold (the false negative rate). Similarly, $l_{i}(\lambda)$ can also represent the proportion of the predicted lesion region $C_{i}(\lambda)$ that does not overlap with the true lesion region $y_{i}^{\ast}$ (ground-truth) when using $(1 - \lambda)$ as the threshold (the false discovery rate).

Our goal is to control the expected loss for a new test data point $x_{n + 1}$, such that:

\begin{equation}
E\left[ l_{n + 1}(\lambda) \right] \leq \alpha.
\end{equation}

Under the assumption that the $n$ calibration data points and the test data point are exchangeable (the joint probability distribution remains unchanged regardless of their order, implying they are independent and identically distributed), the expected loss can be expressed as:

\begin{equation}
E\left[ l_{n + 1}(\lambda) \right] = \frac{nL_{n}(\lambda) + l_{n + 1}(\lambda)}{n + 1}.
\end{equation}

We then seek to find a threshold $\widehat{\lambda}$ in the calibration set that satisfies:

\begin{equation}
\begin{split}
\widehat{\lambda} &= \inf\left\{ \lambda:\frac{nL_{n}(\lambda) + 1}{n + 1} \leq \alpha \right\} \\
&= \inf\left\{ \lambda:L_{n}(\lambda) \leq \frac{\alpha(n + 1) - 1}{n} \right\}.
\end{split}
\end{equation}

With this $\widehat{\lambda}$, we achieve the statistically rigorous guarantee:

\begin{equation}
\begin{split}
E\left[ l_{n + 1}\left( \widehat{\lambda} \right) \right] 
&= \frac{nL_{n}\left( \widehat{\lambda} \right) + l_{n + 1}\left( \widehat{\lambda} \right)}{n + 1} \\
&\leq \frac{nL_{n}\left( \widehat{\lambda} \right) + 1}{n + 1} \leq \alpha.
\end{split}
\end{equation}

This ensures that the expected loss for the test data point is controlled at the user-specified level $\alpha$, providing a statistically rigorous guarantee for the model's performance.

\section{Experiment}

\subsection{Experimental Settings}
Based on the requirements of brain tumor segmentation, this study conducts comparative experiments using two instance segmentation models: Mask R-CNN \cite{ref10}  and BlendMask \cite{ref11}. Mask R-CNN \cite{ref10} , as a classic two-stage method, employs a ResNet50-FPN backbone to generate region proposals and predict masks. BlendMask \cite{ref11}, building upon Mask R-CNN \cite{ref10} , integrates YOLACT's proton network and FOCS feature fusion mechanism to achieve finer instance segmentation. All models adopt a unified ResNet50-FPN backbone. During training, we select appropriate batch sizes and learning rates for each model, along with optimized confidence thresholds and NMS IoU thresholds. We apply random horizontal flipping as the basic data augmentation strategy.

\subsection{Dataset}
Our study utilizes a brain tumor dataset \cite{ref39} comprising 3,064 T1-weighted MRI images, including meningioma (708 images), glioma (1,426 images), and pituitary tumor (930 images). For the purpose of this study, we combine these three tumor types into a single unified category labeled as ``tumor''.

The original dataset stores data in MATLAB format, with each case containing:
\begin{itemize}
    \item Raw image data
    \item Tumor class label
    \item Manually annotated tumor border coordinates
    \item Binary tumor mask
\end{itemize}

For experimental purposes, we randomly partition the dataset \cite{ref39} into:
\begin{itemize}
    \item Training set: 2,040 images (converted to VOC format)
    \item Test set: 1,024 images (used for FDR and FNR evaluation)
\end{itemize}

In our ablation studies, we further divide the test set into calibration and test subsets with equal sizes (512 images each). To ensure statistical robustness, we repeat this division 10 times with different random seeds. Additionally, we conduct ratio experiments by systematically varying the calibration-to-test set ratios from 9:1 to 1:9 (with a single random partition for each ratio).

\subsection{Hyper parameter}
\subsubsection{Mask R-CNN Configuration}
For our Mask R-CNN \cite{ref10}  implementation, we adopt several key hyperparameters optimized for tumor instance segmentation. The model uses a single foreground class (\texttt{NUM\_CLASSES}=1) and trains with a batch size of 8 (\texttt{IMS\_PER\_BATCH}=8) to balance GPU memory constraints and training efficiency. We set the initial learning rate to 0.008 (\texttt{BASE\_LR}=0.008) for single-GPU training, with a total training duration of 26 epochs. The learning rate undergoes reductions by a factor of 0.1 (\texttt{GAMMA}=0.1) at epochs 16 and 22 (\texttt{LR\_STEPS}=[16,22]), following standard practice for this architecture.

\subsubsection{BlendMask Configuration}
In the BlendMask \cite{ref11} implementation, we configure the network with similar design considerations but different optimization parameters. The model processes a single output class (\texttt{NUM\_CLASSES}=1) for tumor segmentation using a batch size of 8 (\texttt{IMS\_PER\_BATCH}=8). We initialize the base learning rate at 0.005 (\texttt{BASE\_LR}=0.005) and train for 9000 iterations (\texttt{max\_iter}=9000), which corresponds to approximately 30 epochs given our dataset size of 2400 images. These parameters reflect empirical optimizations specific to brain tumor segmentation tasks.

\subsubsection{Conformal Prediction Framework}
The conformal prediction framework employs carefully selected parameters to ensure methodological rigor. We systematically evaluate target coverage rates $\alpha$ using $\texttt{np.arange}(0.1, 1.0, 0.1)$ and set candidate thresholds $\lambda$ as $\texttt{np.linspace}(1, 0, 101)$ for precise threshold selection. The experiments include 10 repetitions (\texttt{n\_runs}=10) to assess robustness, with calibration-test split ratios (\texttt{split\_ratio}) varying from 1:1 to 1:9 for comprehensive analysis. Additional operational parameters include a confidence threshold of 0.5 for prediction filtering and a default NMS IoU threshold of 0.5 for redundant box elimination, both chosen through empirical validation.

\subsection{Evaluation Metrics}
\label{subsec:eval_metrics}

We employ the following quantitative measures to assess model performance:

\subsubsection{False Discovery Rate (FDR)}
\begin{equation}
\mathrm{FDR} = 1 - \frac{|\mathcal{P} \cap \mathcal{G}|}{|\mathcal{P}|}
\end{equation}
where $\mathcal{P}$ denotes the predicted mask and $\mathcal{G}$ represents the ground truth. FDR quantifies the proportion of false positive predictions.

\subsubsection{False Negative Rate (FNR)}
\begin{equation}
\mathrm{FNR} = 1 - \frac{|\mathcal{P} \cap \mathcal{G}|}{|\mathcal{G}|}
\end{equation}
FNR measures the fraction of missed tumor regions in ground truth.

\subsubsection{Conformal Prediction Metrics}
For uncertainty quantification, we use:

\begin{itemize}
    \item \textbf{Empirical Coverage Rate (ECR)}: 
    \begin{equation}
    \mathrm{ECR} = \frac{1}{N}\sum_{i=1}^N \mathbb{I}(y_i \in \hat{C}_i)
    \end{equation}
    where $\hat{C}_i$ is the prediction set for sample $i$ and $\alpha$ is the target coverage level.
    
    \item \textbf{Average Prediction Set Size (APSS)}:
    \begin{equation}
    \mathrm{APSS} = \frac{1}{N}\sum_{i=1}^N |\hat{C}_i|
    \end{equation}
    with smaller values indicating higher efficiency.
\end{itemize}

Visual results present error curves comparing empirical FDR/FNR against theoretical bounds, while ablation studies show bar plots of error rates across different data splits (mean $\pm$ std over 10 trials).

\subsection{Theoretical Formulas}

\paragraph*{Theorem 1: FDR Control Formula}
\begin{equation}
\mathbb{P}(\text{FDR}(X_{n+1}) \leq \alpha) \geq 1 - \delta.
\end{equation}

\paragraph*{Theorem 2: Threshold Selection Formula}
\begin{equation}
\widehat{\lambda} = \inf\left\{\lambda : \frac{1}{n}\sum_{i=1}^n L_i(\lambda) \leq \alpha - \frac{B-\alpha}{n}\right\}.
\end{equation}

\subsection{Practical Steps}

\subsubsection*{Experiment 1 (Precision)}
\begin{enumerate}
\item Parameter selection: Calibration loss $L_i(\lambda)$ as:
\begin{equation}
l_i(\lambda) = 1 - \frac{|C_i(\lambda) \cap V_i^*|}{|C_i(\lambda)|}, \quad B=1,
\end{equation}
set $\alpha$ from 0.1 to 0.9 in steps of 0.1.
\item Calibrate threshold using threshold selection formula on calibration set.
\end{enumerate}

\begin{figure}[htbp]
    \centering
    \includegraphics[width=0.5\textwidth]{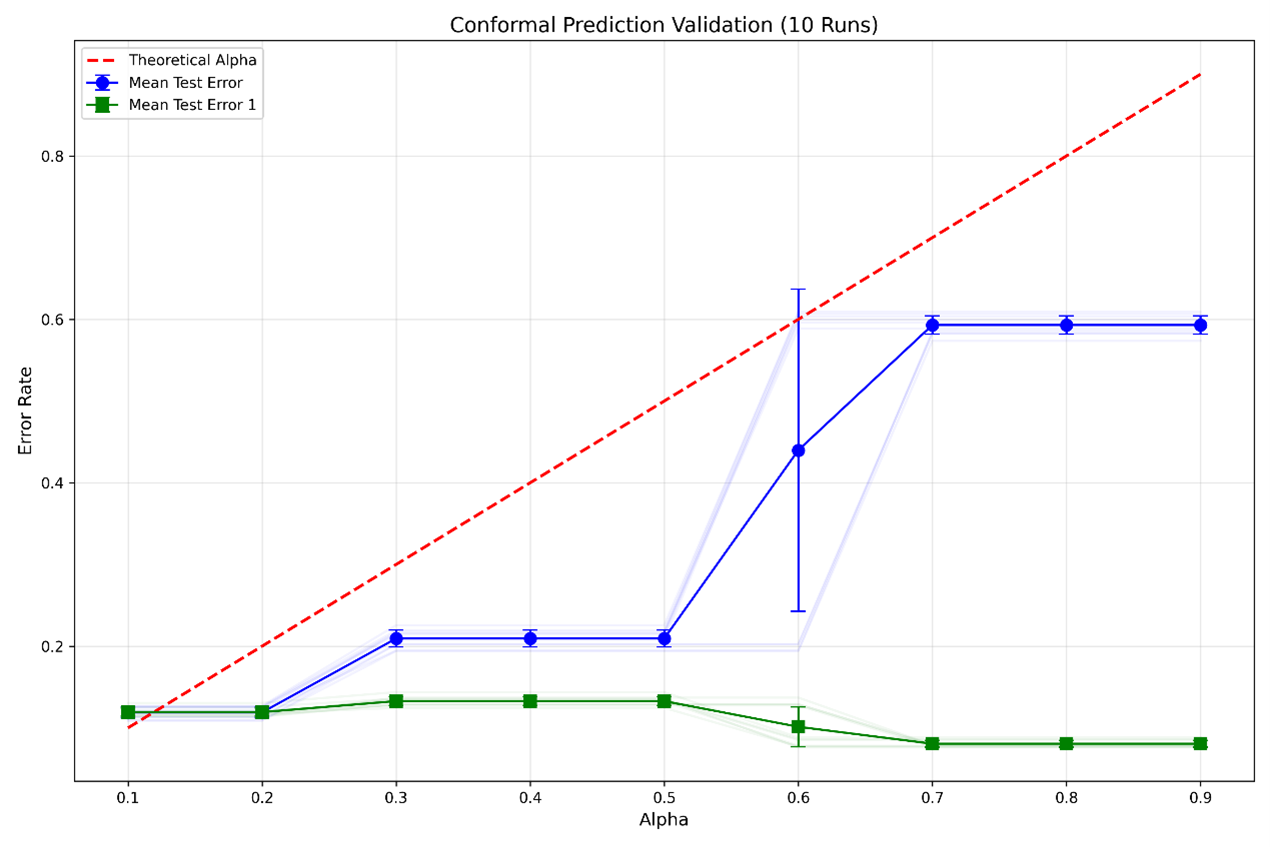}
    \caption{Visualization of Mask R-CNN FDR results. Red dashed line represents $\alpha$, blue line represents mean FDR loss, green line represents FNR loss.}
    \label{fig:maskrcnn_fdr}
\end{figure}

\begin{figure}[htbp]
    \centering
    \includegraphics[width=0.5\textwidth]{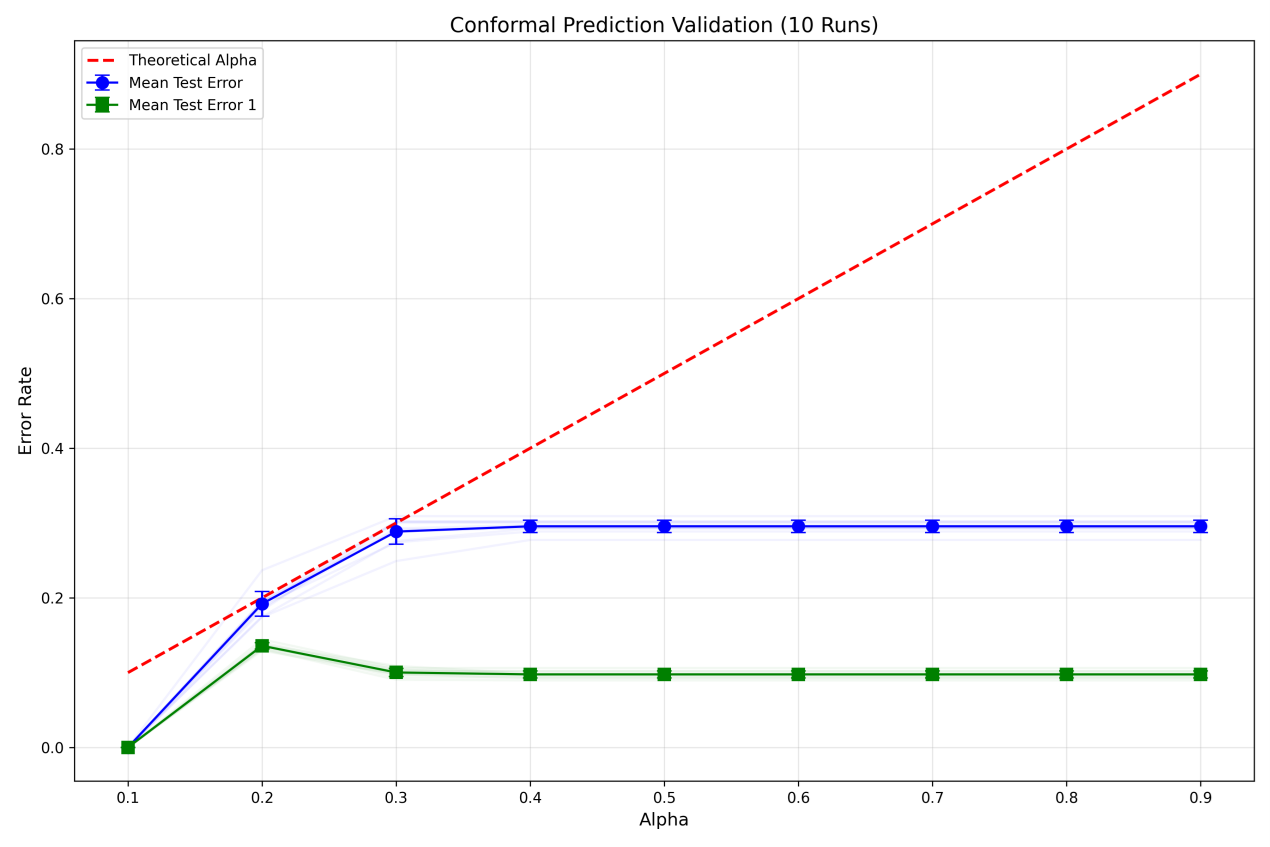}
    \caption{Visualization of BlendMask FDR results. Red dashed line represents $\alpha$, blue line represents mean FDR loss, green line represents FNR loss.}
    \label{fig:BlendMask_fdr}
\end{figure}

In the experiment, we select \textbf{FDR} as the core metric for the calibration loss function, focusing on the precision of segmentation results. This choice ensures that the output of the prediction model remains as accurate as possible, avoiding excessive false positives (false discoveries), even if it might miss some target regions. We use both Mask R-CNN \cite{ref10}  and BlendMask \cite{ref11} models and generate 10 test results through cross-validation.

As shown in Fig. \ref{fig:maskrcnn_fdr} and Fig. \ref{fig:BlendMask_fdr}, the blue curve shows the trend of FDR, with its mean consistently remaining below the red dashed line at $\alpha=0.25$. The blue shaded area indicates the fluctuation range of FDR, demonstrating the stability of the method. The green shaded area represents the fluctuation range of FNR, showing that the method achieves a balance between precision and coverage. These results demonstrate that our method satisfies the user-specified risk level.

\subsubsection*{Experiment 2 (Coverage)}
\begin{enumerate}
\item Parameter selection: Calibration loss $L_i(\lambda)$ as:
\begin{equation}
l_i(\lambda) = 1 - \frac{|C_i(\lambda) \cap V_i^*|}{|V_i^*|}, \quad B=1,
\end{equation}
set $\alpha$ from 0.1 to 0.9 in steps of 0.1.
\item Calibrate threshold using threshold selection formula on calibration set.
\end{enumerate}

\begin{figure}[htbp]
    \centering
    \includegraphics[width=0.5\textwidth]{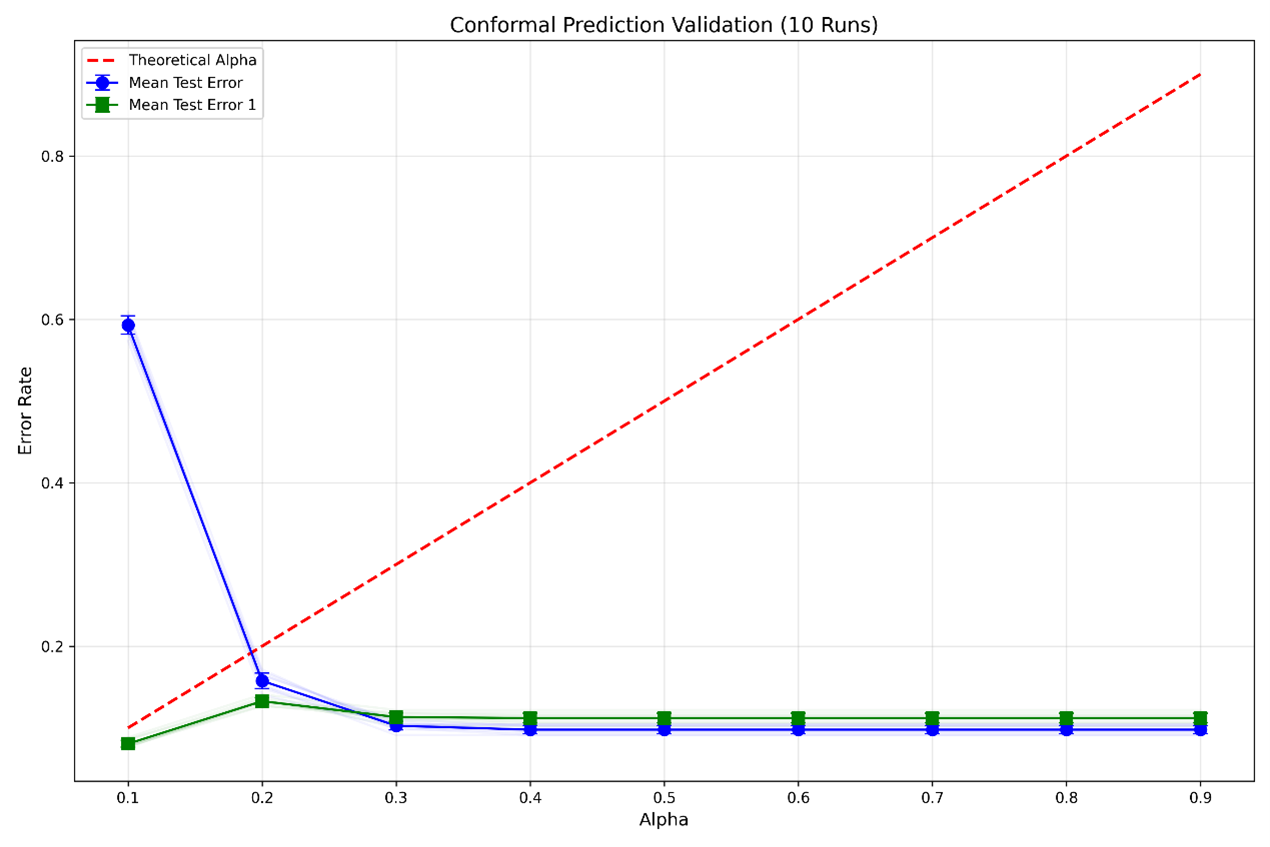}
    \caption{Visualization of Mask R-CNN FNR results. Red dashed line represents $\alpha$, blue line represents mean FNR loss, green line represents FDR loss.}
    \label{fig:maskrcnn_fnr}
\end{figure}

\begin{figure}[htbp]
    \centering
    \includegraphics[width=0.5\textwidth]{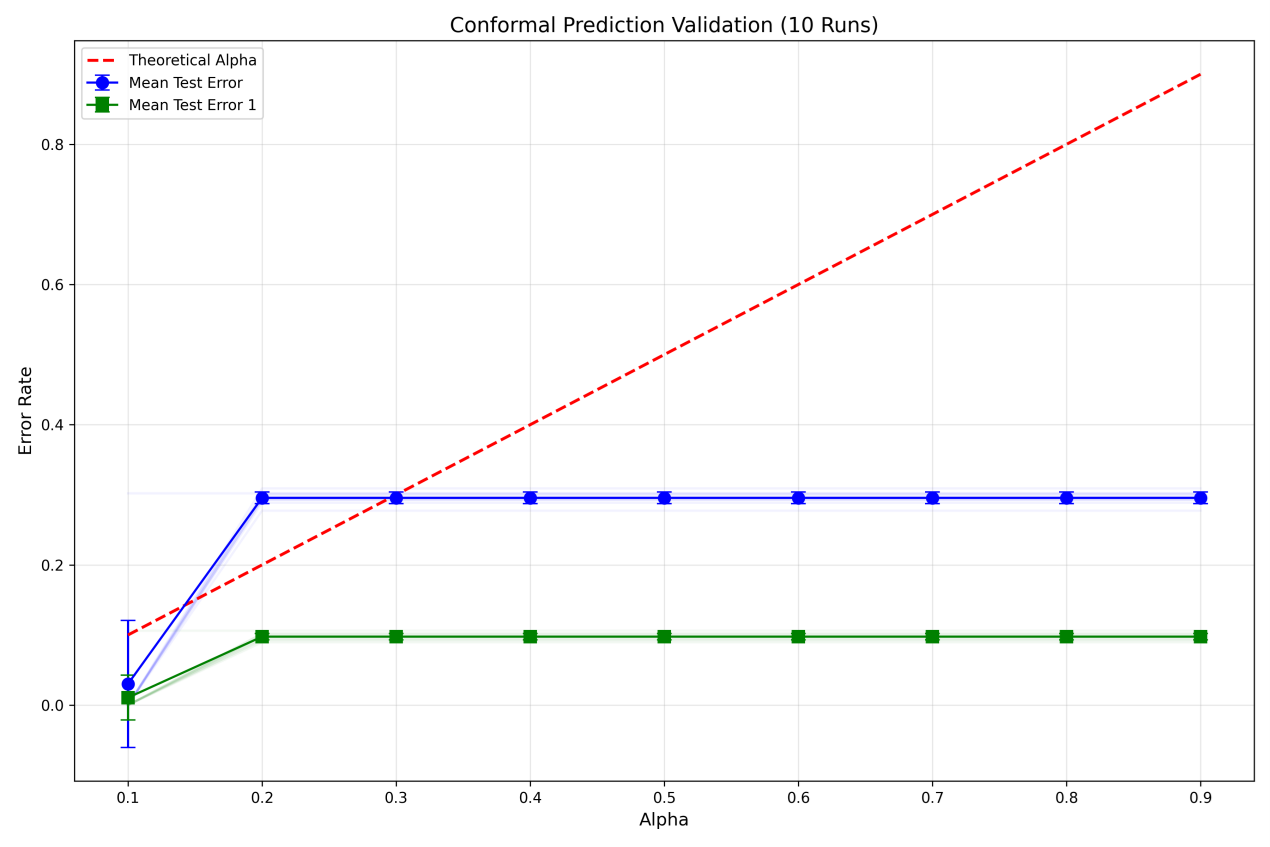}
    \caption{Visualization of BlendMask FNR results. Red dashed line represents $\alpha$, blue line represents mean FNR loss, green line represents FDR loss.}
    \label{fig:BlendMask_fnr}
\end{figure}

In the experimental design, we focus on \textbf{FNR} as the primary goal of the calibration loss function, aiming to maximize the coverage of segmentation results, i.e., covering as many correct tumor regions as possible. This choice ensures that the prediction model captures all possible target regions comprehensively, even at the cost of increasing some false positives (false negatives). To verify the effectiveness of this method, we use both Mask R-CNN \cite{ref10}  and BlendMask \cite{ref11} models and generate 10 test results through cross-validation.

As shown in Fig. \ref{fig:maskrcnn_fnr} and Fig. \ref{fig:BlendMask_fnr}, the green curve shows the trend of FNR, while the blue curve shows the trend of FDR, and the red dashed line indicates the set significance level $\alpha=0.25$. As the precision requirement changes, the performance of FNR under different thresholds shows some fluctuations but remains within the $\alpha$ range overall. Although higher precision leads to a higher loss in accuracy in some cases, the results demonstrate that our method effectively controls FNR within the user-specified risk level while ensuring comprehensive coverage of the segmentation results.

\subsubsection*{Experiment 3 (Different Calibration-to-Test Set Ratios)}
We evaluate the final threshold on the prediction set, visualize the prediction loss (including FDR and FNR) and loss upper bound ($\alpha=0.1$ to 0.9), and verify whether the results satisfy the control formula.

\begin{figure}[htbp]
    \centering
    \includegraphics[width=0.5\textwidth]{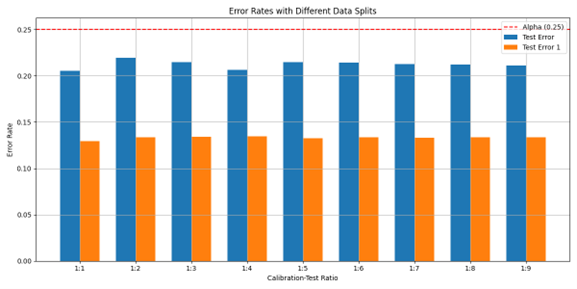}
    \caption{Ablation study results for Mask R-CNN. Blue line represents FDR loss, orange line represents FNR loss, red dashed line represents fixed $\alpha=0.25$.}
    \label{fig:maskrcnn_ablation}
\end{figure}

\begin{figure}[htbp]
    \centering
    \includegraphics[width=0.5\textwidth]{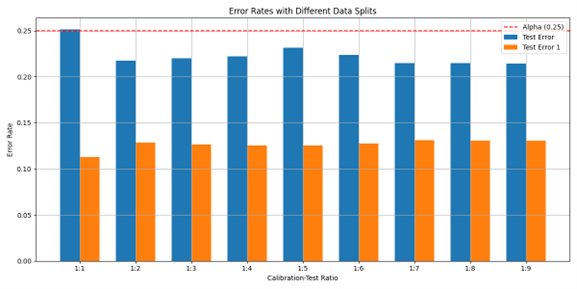}
    \caption{Ablation study results for BlendMask. Blue line represents FDR loss, orange line represents FNR loss, red dashed line represents fixed $\alpha=0.25$.}
    \label{fig:BlendMask_ablation}
\end{figure}

To evaluate the robustness of the model under different calibration-to-test set ratios, we conduct an ablation study. This experiment verifies whether our method can still provide statistical guarantees at the user-specified risk level, even when the ratio of calibration to test sets varies. We adjust the ratio of the calibration set to the test set from 1:1 to 1:9 and record the FDR and FNR losses under each ratio.

As shown in Fig. \ref{fig:maskrcnn_ablation} and Fig. \ref{fig:BlendMask_ablation},The experimental results show that the blue curve represents FDR loss, the orange curve represents FNR loss, and the red dashed line still indicates the set significance level $\alpha=0.25$. Regardless of how the ratio of the calibration set to the test set changes, the losses of FDR and FNR remain controlled within the $\alpha$ range, with relatively small fluctuations. In particular, when the proportion of the calibration set increases, the means of both FDR and FNR decrease, demonstrating the positive role of more calibration data in adjusting the threshold. The ablation study results show that our method maintains strong robustness and achieves effective error rate control under different data distribution conditions.

\section{Conclusion}
In this work, we present a novel approach to brain tumor segmentation by integrating conformal prediction with advanced deep learning models. Our method addresses a critical challenge in medical image analysis: ensuring statistical guarantees for segmentation performance regardless of the underlying model's capabilities.

We adapt the conformal prediction framework to the specific context of brain tumor segmentation by designing tailored loss functions that represent either false discovery rate (FDR) or false negative rate (FNR), allowing users to prioritize either precision or recall based on clinical requirements. By leveraging a small exchangeable calibration set, our approach dynamically adjusts the segmentation threshold to satisfy user-specified risk levels.

Future work includes:
\begin{itemize}
\item Extending this approach to multi-instance segmentation tasks
\item Investigating its applicability to other medical imaging modalities beyond brain tumors
\end{itemize}

\bibliographystyle{unsrtnat}
\bibliography{references}

\end{document}